\def\year{2018}\relax
\documentclass[letterpaper]{article} 
\usepackage{aaai18}  
\usepackage{times}  
\usepackage{helvet}  
\usepackage{courier}  
\usepackage{url}  
\usepackage{graphicx}  
\usepackage{epsfig} 
\usepackage{enumerate,graphicx,epsfig,subfigure}
\usepackage{amsmath,amssymb,amsfonts,dsfont}
\usepackage{psfrag}
\usepackage{tikz-cd}

\usepackage{multirow}  
\usepackage{makecell}

\renewcommand{\th}{\theta}

\newcommand{\BBR}{\mathbb{R}}

\newcommand{\CALB}{\mathcal{B}}
\newcommand{\CALC}{\mathcal{C}}
\newcommand{\CALD}{\mathcal{D}}

\newcommand{\CALF}{\mathcal{F}}

\newcommand{\CALS}{\mathcal{S}}

\newcommand{\CALR}{\mathcal{R}}

\newcommand{\set}[2]{\{#1\colon#2\}}

\newcommand{\norm}[1]{\left\Vert#1\right\Vert}

\newcommand{\ind}[1]{{\bf 1}_{#1}}

\DeclareMathOperator*{\argmin}{\arg\hspace{-0.07em}\min}

\nocopyright

\begin{document}
%
\title{
    Human Understandable Explanation Extraction for Black-box Classification Models
    Based on Matrix Factorization
}

\author{Jaedeok Kim and Jingoo Seo\\
Artificial Intelligence Team, Software R\&D Center, Samsung Electronics\\
56 Seongchon-gil, Seocho-gu, Seoul, Republic of Korea\\
}

\maketitle
\begin{abstract}
    In recent years, a number of artificial intelligent services have been developed such as defect detection system or diagnosis system for customer services.
    Unfortunately, the core in these services is a black-box in which human cannot understand the underlying decision making logic, even though the inspection of the logic is crucial before launching a commercial service.
    Our goal in this paper is to propose an analytic method of a model explanation that is applicable to general classification models.
    To this end, we introduce the concept of a contribution matrix and an explanation embedding in a constraint space by using a matrix factorization.
    We extract a rule-like model explanation from the contribution matrix with the help of the nonnegative matrix factorization.
    To validate our method, the experiment results provide with open datasets as well as an industry dataset of a LTE network diagnosis and the results show our method extracts reasonable explanations.
\end{abstract}

\section{INTRODUCTION}

    In recent years, a number of artificial intelligent services have been developed such as defect detection system or diagnosis system for customer services.
    The core in these services consists of a classification model learned from historical data that classified by domain experts.
    To have a better model many advanced techniques have been applied like ensemble, boosting, or deep learning, and these techniques have helped in improving the performance of a classification model \cite{kotsiantis2007supervised}.
    Unfortunately, such model has a black-box problem and human cannot understand how the model makes a decision.
    The black-box model has difficulty to give a trust for users even if the model shows a low classification error.

    This is a problem of a model inspection.
    From the viewpoint of service providers, the inspection of the model behavior is crucial before launching their commercial services.
    Since service providers have potentially risks in applying a black-box model, they spend much time to verify a model until its reliability is proved and it increases development costs significantly.
    So the service providers have still preferred a hybrid of a machine learning based model and a rule based model in their services.

    We consider the model inspection problem of a classification model.
    To solve this problem, we propose a method to extract rule-like explanations of a classification model to account for the underlying decision making constraint.
    Our approach can help a service provider understand the decision making logic of the model and quickly decide whether the model is acceptable or not.
    As a consequence, our result will help significantly reduce the efforts in the model inspection.

    Explaining a black box model is not an easy task in general because of the variety of algorithms to build a classification model.
    There have few works to analyze how a model predicts a result for a given instance.
    To understand the reason of a classification result in an image classification problem, visualization techniques such as the saliency map have been used \cite{SimonyanVZ13,ZintgrafCAW17}.
    Similar to the silence map, a trust problem on a prediction result has been tackled \cite{ribeiro2016should}.
    Their approach can explain the reason of prediction results by a model in the instance level,
    but it is not enough to inspect the characteristics of a model.

    Our method is based on their results.
    We will extract common decision constraints from explanations of classification results by using the nonnegative matrix factorization.
    When we have a set of explanations of a classification result, the explanations shares the same characteristics if the model has underlying decision making constraints.
    For instance, when experts classify system diagnosis issues into several categories, they have their own decision making constraints even if they cannot recognize it explicitly.
    The constraints should be reflected in the explanations of instances.

    The main contribution in this paper is that we propose an analytic method of a model explanation that is applicable to general classification models.
    We introduce the concept of a contribution matrix and an explanation embedding in a constraint space by using the matrix factorization.
    We experimentally show that the embedded explanations are well clustered according to the classification category of instances.
    By using the explanation embedding, we extract a model explanation of the rule-like form.
    We also perform experiments with open datasets as well as an industrial dataset to show the validity of our approach in practice.
    In particular, many of industrial datasets, especially defect detection or diagnosis systems, consist of numerical attributes each of them has its own meaning.
    Our method therefore will focus on dataset having numeric attributes.


\section{Related Works}

    Understanding a model is one of the key issues in the machine learning field.
    However, understanding a black-box model is difficult in general because of the variety of algorithms and the dependency of data characteristics.
    So the decision tree or the support vector machine have been still popular and their variations have been published over decades because those algorithms have intuitive structures to be interpreted \cite{witten2016data}.
    Unfortunately, they do not guarantee a sufficiently good performance in practice and we often need to apply advanced techniques to improve the performance, such as ensemble or boosting, which makes a model be a black box like a deep neural network that is the promising technique nowadays.

    Visualization techniques such as the saliency map have been often used in order to understand classification results in the image classification problem.
    To visualize the evidences of an image classification model, \cite{SimonyanVZ13} measures the sensitivity of classifications and \cite{ZintgrafCAW17} analyzes the activation difference in the marginal distribution of a small area.
    These approaches are specified to image classifiers, while they are not easy to be extended to the other applications.

    To overcome the difficulty of model understanding, the explanation of a black box model has been studied in recent years.
    Lipton discussed the meaning of interpretability and the properties of interpretable models \cite{lipton2016mythos}.
    Tuner introduced several axioms to define the concept of explanation such as the {\it eligibility} of explanation in terms of the conditional probability \cite{Turner2016}.
    A model explanation system was also developed to explain results of a black box model based on the axioms.
    Riberio et al. tackled the trust problem and proposed a method to explain a classification result \cite{ribeiro2016should}.
    The key idea is that a classification model can be locally approximated by perturbing each instance and the approximation can explain how much each feature affects the decision.
    The local perturbation is also used in \cite{fong2017interpretable} for interpreting prediction results in the image classification domain.
    In \cite{ShrikumarGK17}, Shrikumar et al. tackled such issue and proposed a method to determine the important features by backpropagating the contributions of neurons.
    Ross et al. develop a method to train a model with the {\it right reasons} \cite{RossHD17}.
    For a differentiable classifier such as neural network, they considered the first order derivative of the classifier as an explanation and designed a loss function with a consideration of an explanation.
    In the above literatures, only instance level explanation has been focused and a model level explanation has been overlooked.

    A model level explanation necessarily needs to understand the underlying decision making constraint of a given classification model.
	Kim et al. proposed a simplification method of Bayesian network model \cite{kim2014bayesian}.
	Rule extraction methods have been proposed in \cite{nunez2002rule}, but they are restricted in the support vector machine.
    Bastani et al. try to understand the reasoning process of a black box model by extracting a decision tree that approximates the classification model \cite{bastani2017interpreting}.

\section{Explanation Modeling}

	In this section, we would develop a mathematical model to describe precisely a model explanation.

    A classification model is a function $f\colon\CALS\to\CALC$ where $\CALS$ is a space of instances and $\CALC=\{1,\cdots,C\}$ is a category space.
    Since we want to focus on a dataset consisting of only numerical attributes without any categorical attribute, we suppose $\CALS$ is in $\BBR^m$ where $m$ is the number of attributes.
    Let $\CALD$ be a subset of $\CALS$ of size $N$, called a dataset.
    Let $R_{c}(\CALD)$ be the set of instances in a dataset $\CALD \subset \CALS$ that are classified by $c$, i.e.,
    \begin{align*}
    	R_c(\CALD) := \set{x \in \CALD}{f(x)=c}.
    \end{align*}
    Also, we let $\CALR(\CALD):=\{R_1(\CALD), \cdots, R_C(\CALD)\}$ be a partition of $\CALD$.
    For convenience, we will use $R_{c}$ instead of $R_{c}(\CALD)$ unless there is ambiguity.

    The purpose of a model explanation is to understand classification constraints for each category.
    For numerical attributes the easiest way to intuitively understand a constraint is checking whether a value of an attribute is in a certain interval or not.
    So it is enough to find rough constraints to approximate each of the partition $\CALR$, which can be done as the following.
    \begin{align*}
    	R_c & \approx \bigcup_{i} B_{c,i} =: B_c
    \end{align*}
    where $B_{c,i}$ is a subset of the form
    \begin{align*}
    	B_{c,i}
        & := \set{(x_1,\cdots,x_m)\in\CALS}{a_j< x_j\le b_j,j=1,\cdots,m} \\
        &  = \prod_{j=1}^m [a_j < x_j \le b_j],
    \end{align*}
    called a rectangle.
    Here, $a_j$ or $b_j$ can have the infinity values $-\infty$ or $\infty$ which induces a half bounded interval or a unbounded interval.

    We then define a model explanation of $f$ by a collection
    \begin{align}
        \label{def:ModelExp}
        \CALB:=\{B_1,\cdots,B_C\}
    \end{align}
    of the set $B_c$ of rectangles.
    In addition, we call $B_c$ an model explanation of the category $c$.

    There is an advantage of approximating each $R_c$ instead of approximating the function $f$.
    Each component $B_{c,i}$ of $B_c$ is a product of intervals.
    So the component $B_{c,i}$ is directly matched with a rule-like explanation which is very simple to understand.
    In the existing literatures, a decision tree has been considered as an interpretable model and algorithms to extract a decision tree from a classification model have been developed \cite{bastani2017interpreting}.
    A decision tree may have an advantage of understanding reasoning process since in the decision tree a path from the root to a leaf intuitively represents a process of decision making.
    However, a decision tree is still hard to understand as a simple rule because of the instability of a tree structure \cite{Li2002}.

    Extracting an approximation model $f^{(a)}$ is about finding {\it a hard constraint} on decision boundaries.
    In this approach, every instance has one category, that is, $(f^{(a)})^{-1}(c):=\set{x\in\CALS}{f^{(a)}(x)=c}$, $c\in\CALC$, are mutually disjoint.
    To do this, $f^{(a)}$ should decide which category is more suitable for an instance even if it lies on a boundary between the areas of two different categories.
    It follows that $f^{(a)}$ needs more complex constraints as the complexity of $f^{(a)}$ increases.
    Consequently, it leads barely understandable explanations although $f^{(a)}$ approximates accurately.
    However, our approach, extracting $\CALB$, is about finding {\it a soft constraint} since $B_c$ is allowed to be overlapped.
    The complexity of an explanation needs not to increase as much as that of $\hat{f}$ increases.
    In fact, from the viewpoint of extracting understandable explanations, it is unnecessary to have a very accurate approximation of $f$.


\subsection{Performance Measure}

    In order to measure the quality of a model explanation $\CALB$, we use the $F_1$ score.
    Note that a model explanation $B_c$ of a category $c$ induces a binary classification $\hat{f}_c$ such that $\hat{f}_c(x) = c$ if $x\in B_c$.
    Consider a restriction $f_c$ of $f$ to a binary classification, i.e., $c$ and others.
    With considering a classification result by $f_c$ as a true value, we use a relative accuracy of $\hat{f}_c$ to $f_c$.
    Then the performance of $B_c$ over a dataset $\CALD$ is defined by  the $F_1$ score of $\hat{f}_c$ relative to $f_c$ over the dataset $\CALD$
    \begin{align*}
        F_1(B_c) := F_1(f_c, \hat{f}_c).
    \end{align*}
    Finally, the performance of a model explanation is defined by
    \begin{align}
        \label{def:Perf}
        F_1(\CALB) := \frac{1}{C} \sum_{c\in\CALC} F_1(B_c).
    \end{align}

    Note that there is no dependency between $F_1(B_c)$ and $F_1(B_{c'})$ for $c\neq c'$.
    The maximization of $F_1(\CALB)$ is achievable by solving subproblems each of which maximizes the value of $F_1(B_c)$ for each category $c$.
    So we decompose our problem of finding $\CALB$ into $c$ subproblems.
    Therefore, we tag a category $c$ as our target category and then assume $f$ is a binary classification model of two labels, say c (the target category) and 0 (others), through the following sections.

\section{Model Explanation Extraction}

    \begin{figure}[t]
        \centerline{\includegraphics[width=\columnwidth]{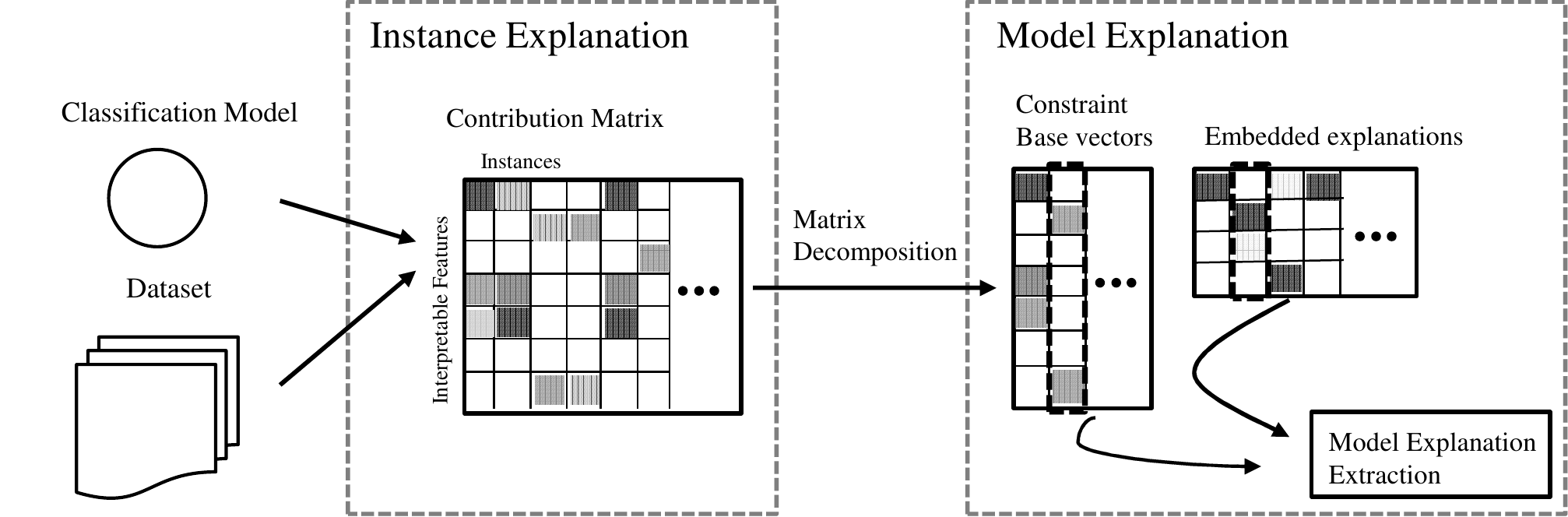}}
        \caption{
            Overview of a model explanation extraction method.
        }
        \label{fig:flow}
    \end{figure}

    Extracting a model explanation consists of two main steps.
    Figure \ref{fig:flow} summarizes the flow of our method.
    At first, an explanation for each instance will be generated by measuring how much an attribute of the instance contributes a classification result.
    The next step will construct a contribution matrix and then extract the common behaviors in the contribution matrix with the help of nonnegative matrix factorization.
    The detail of each step will be explained in the following subsections.

\subsection{Interpretable Instance Explanation}

    To understand the behavior of a classification model, we first need to explain which attributes of an instance mainly affect a result and measure the amount of their impacts in a proper way.
    To this end, we consider a local perturbation of $f$ at a given instance $x$.
    In this paper, the perturbation is applied to interpretable features instead of attributes of an instance borrowing the idea introduced in \cite{ribeiro2016should}.
    The advantage of this approach is that the complexity of explanations is controllable by restricting the number of interpretable features.

    The problem of finding an instance explanation can be formulated as in the follows.
    We assume that $f$ is decomposed by an embedding $\varphi$ to the interpretable space $\CALF$ and an interpretable model $\tilde{f}\colon\CALF\to\CALC$ satisfying $f = \tilde{f} \circ \varphi$.
    \begin{equation*}
    	\begin{tikzcd}[row sep=2em, column sep=3em]
          \CALS \arrow[d,"\varphi"] \arrow[r,"f"] & \CALC \\
          \CALF \arrow[ru,shift right,"\tilde{f}"]
        \end{tikzcd}
    \end{equation*}

    Let $\CALF$ be an interpretable feature space of finite dimension whose dimension represents an interpretable feature.
    For numerical attributes, we may consider an interval of attribute value as an interpretable feature explained in the previous section.
    Since we are able to deal with only finite number of features, the domain of attributes should be discretized.

    Recalling that a rectangle has the form of a product of intervals, we will use an indicator of the form $\ind{[x_i\le b]}$ as an interpretable feature.
    Only upper bounded interval is enough since $\ind{[x_i\le b]}(x)=0$ implies $x_i>b$.
    Now we denote by $F$ the set of indicators of a half-bounded interval.
    Then $\CALF=\{0, 1\}^{M}$ is the interpretable space where $M:=|F|$.
    That is, for an element $z=(z_1,\cdots,z_M)=\varphi(x)$ in $\CALF$, $z_i=1$ implies $\ind{[x_{j_i}\le b_i]}(x)=1$ for the corresponding indices $j_i$ and $b_i$.

    Each feature will be directly used to classify categories without any further computation for the understandability of $\tilde{f}$.
    So we restrict $\tilde{f}\colon\CALF\to\CALC$ to a binary classification model whose decision boundary is given by a linear function $z\mapsto\phi^T z$ where $\phi=(\phi_0, \cdots, \phi_{M-1})^T$, that is, $z$ is classified by the target category $c$ if and only if $\tilde{f}(z) > 0$.
    Then to find a suitable $\tilde{f}$ for a given instance $x$ an optimization problem can be formulated as the following.
    \begin{align}
        \label{eqn:OptProblem}
        \tilde{f}^* = \argmin_{\tilde{f}} J_x(f, \tilde{f})
    \end{align}
    Here $J_x$ is a distance function between two classification models $f$ and $\tilde{f}$ over a perturbation set of $x$.
    If we let ${\phi_f^*}^T z$ be the decision boundary of $\tilde{f}^*(z)$, its normal vector $\phi_f^*=(\phi^*_{f,1},\cdots,\phi^*_{f,M})^T$ is said to be an explanation of the instance $x$.
    We call $\phi^*_{f,i}$ the contribution of the $i$th feature on $x$ with respect to $f$.
	If $\phi^*_{f,i} > 0$, the $j$th feature $[x_{j_i}\le b_i]$ positively affects the decision of $x$.
	Otherwise, $\phi^*_{f,i} < 0$ implies the $i$th feature negatively affects.
    In the later section, with the help of LIME \cite{ribeiro2016should}, we will evaluate $\phi_f^*$ by defining $J_x$ with the $F_1$ score that represents the relative difference between the results of $f$ and those of $\tilde{f}$.
	
    Moreover, collecting all instance explanation we define the contribution matrix of a model $f$ with respect to a set $\CALD$ by
    \begin{align}
        \label{eqn:ContrMatix}
        \Phi_f(\CALD) :=
        \begin{bmatrix}
            \phi_{1,1}& \cdots & \phi_{1,N}\\
            \vdots & \ddots & \vdots \\
            \phi_{M,1} & \cdots & \phi_{M,N}
        \end{bmatrix}.
    \end{align}
    Here, $\phi_j = (\phi_{j,1},\cdots,\phi_{j,M})^T$ is the instance explanation of $x_j\in\CALD$, i.e., the solution of (\ref{eqn:OptProblem}).
    For convenience an contribution matrix will be denoted by $\Phi_f$ instead of $\Phi_f(\CALD)$ if there is no ambiguity.

\subsection{Model Explanation from Contribution Matrix}

    The key intuition to extract a model explanation $\CALB$ is the contribution matrix $\Phi_f$ contains the underlying constraints of decision making.
    By definition of $\Phi_f$, the $i$th row of the matrix $\Phi_f$ is the contributions of the $i$th feature.
    So the column vectors of $\Phi_f$ will be grouped into several clusters according to the decision making constraints of the classification model $f$, which can be found by using a matrix factorization.
    In particular, the nonnegative matrix factorization (NMF) can find base vectors whose elements are positive and represent how much a feature affects.
    As a consequence, rule-like explanations can be achievable by interpreting those base vectors.


    The contribution matrix $\Phi_f$ contains both positive and negative values while we need a nonnegative matrix to apply the NMF.
    Let us consider a decomposition $\Phi_f=\Phi^+-\Phi^-$ where $\Phi^+=[\phi^+_{i,j}]_{i,j}$ is the positive matrix of $\Phi$ defined by $\phi^+_{i,j}=\phi_{i,j} \ind{\phi_{i,j} > 0}$.
    Analogously, $\Phi^-$ is the negative matrix of $\Phi$.
    Instead $\Phi_f$, we consider a transformed contribution matrix $\bar{\Phi}_f$ of dimension $2M \times N$,
    \begin{align*}
        \bar{\Phi}_f :=
            \begin{bmatrix}
                \Phi^+ \\ \Phi^-
            \end{bmatrix},
    \end{align*}
    whose elements are all nonnegative.
    $\phi^-_{i,j}$ is the negative contribution of the $i$th feature $z_i=\ind{[x_{j_i}\le b_i]}$ to the $j$th instance, which means the $x_{j_i}$ is more preferable to be in $[x_{j_i} > b_i]=[x_{j_i} \le b_i]^c$.
    We take the complementary feature $z'_i =1 - \ind{[x_{j_i}\le b_i]}$ of $z_i$ as the corresponding feature of the $i$th row of $\Phi^-$.
    We thus have $2M$ features.

    We now consider the NMF of the nonnegative matrix $\bar{\Phi}$.
    Let $k$ be an integer such that $k\le 2M, N$ and $W$ and $H$ be nonnegative matrices of dimension $2M\times k$ and $k\times N$, respectively, such that $W$ and $H$ are the solution of
    \begin{align*}
    	\min_{W, H} & \norm{\bar{\Phi}_f - WH}.
    \end{align*}
    We then have
    \begin{align}
    	\label{eqn:expl_decomp}
    	\phi_j \approx W h_j = \sum_{l=1}^k h_{l,j} w_l
    \end{align}
    where $h_j=(h_{1,j},\cdots,h_{k,j})^T$ is the $j$th column vector of $H$ and $w_l=(w_{1,l},\cdots,w_{2M,l})^T$ is the $l$th column vector of $W$.

    A vector $w_l$ can be interpreted as a underlying decision making constraint of the classification model $f$.
    The value of an element $\phi_{i,j}$ in $\bar{\Phi}_f$ represents the relationship between the $i$th feature and the classification result of the $j$th instance.
    Each $w_l$ is a base vector of the contribution matrix $\Phi_f$.
    So $w_l$ contains information of the feature combination that simultaneously affects the decision making of a category.
    Hence, we call the column space of $W$ a constraint space and $w_l$ a base vector of the constraint space.

    Consider a base vector $w_l$.
    Let $\th_w$ be a threshold that filters out meaningless features from the base vector $w_l$.
    We then have constraints by taking every feature that satisfies $w_{i,l} > \th_w$.
    For instance, if $w_{i,l}>\th_w$ and $1\le i\le M$, we will take the corresponding constraint $[x_{j_i}\le b_i]$, and if $M<i\le 2M$, $[x_{j_{i-M}} > b_{i-M}]$ will be taken.
    Combining all constraints we obtain a rectangle of a underlying decision making constraint
    \begin{align*}
        B(w_l, \th_w)
        & := \prod_{i=1}^M [x_{j_i}\le b_i]^{\ind{[w_{i,l} > \th_w]}}
                           [x_{j_i}  > b_i]^{\ind{[w_{i+M,l} > \th_w]}}.
    \end{align*}
    For convenience $A^0$, $A\subset\CALS$, denotes the entire space $\CALS$ of instances.
    If the value of $\th_w$ becomes smaller, $B(w_l, \th_w)$ takes the more constraints and the rule will be more complex.
    On the other hand, a larger value $\th_w$ yields a less complex constraint $B(w_l, \th_w)$.

    A column vector $h_j=(h_{1,j},\cdots,h_{k,j})^T$ of $H$ can be geometrically interpreted as a projection of $\phi_j$ onto the constraint space.
    Specifically, $h_{l,j}$ is the effect of the $l$th base vector on the $j$th instance.
    We call the column vector $h_j$ an embedded explanation.
    If there is a common decision making constraint in classification, the same rule should be applied to determine a category.
    Hence, the embedded explanations are expected to be concentrated in several clusters and almost all vectors in each cluster are labeled by the same category.
    In the later section, we will validate this assumption with experimental results.

    Suppose there are $r$ clusters of embedded explanations and denote by $G_1,\cdots,G_r$.
    Since we have focused on a binary classification, the most frequent label in a cluster is $c$ (the target category) or 0 (the others).
    Let $G_1,\cdots,G_{r_0}$ be the clusters in which elements are mostly labeled by the target category $c$.
    Let a vector $g_i =(g_{i,1},\cdots,g_{i,k})$ be a centroid of $G_i$.
    Then the vector $g_i$ is the representative of $G_i$.
    This can be interpreted as follows.
    In $G_i$, $i\le r_0$, the $l$th base vector $w_l$ contributes to the category $c$ with weight $g_{i,l}$ in average.
    Suppose an instance $x_j$ is classified by the target category $c$ and the corresponding embedded explanation $h_j$ is in the cluster $G_i$.
    If we take the largest element of $g_i$, say $g_{i,1}$, the corresponding base vector $w_1$ is related to the most important explanation of $x$ and therefore the corresponding constraint $B(w_1,\th_w)$ is the most significant for classifying $x_j$ as the target category $c$ and its weight is given by $g_{i,1}$.
    Similarly, the second largest element of $g_i$, say $g_{i,2}$, yields the next important constraint $B(w_2,\th_w)$ with weight $g_{i,2}$.

    From these observations, we extract an explanation of the target category $c$ from each cluster.
    Consider a threshold $k_{\th}$ to take important base vectors that affect significantly.
    We then take top $k_{\th}$ base vectors and denote the set of these vectors by $S_i$.
    By combining all corresponding conditions of base vectors from $S_i$, we have a rectangle
    \begin{align*}
        B_{c,i} = \bigcap_{l\in S_i} B(w_l, \th_w),
    \end{align*}
    which is related to a underlying decision making constraint of $G_i$.
    Since there are $r_0$ clusters relevant to the target category $c$,
    by collecting all constraints we finally obtain a model explanation of the target category $c$
    \begin{align*}
        B_c = \bigcup_{i=1}^{r_0} B_{c,i}.
    \end{align*}

    Summarizing all the above observations, we formulate an optimization problem to find a model explanation of the target category $c$.
    \begin{align}
        \label{eqn:OptModelExp}
        \max_{r, \th_w, k_{\th}} & \quad F_1(B_c), \\
        s.t. & \quad r \le r_{max.}, \ k_{\th} \le k \nonumber
    \end{align}
    The complexity of $B_c$ depends on the value of $r_0$ which is bounded by the value of $r$.
    For human understandable explanation, we consider a constraint $r_{max}$ of the number $r$ of clusters in our optimization problem as in (\ref{eqn:OptModelExp}).
    By solving (\ref{eqn:OptModelExp}) we have a model explanation of the target category $c$.

    Solving the optimization problem for all categories we finally have a method to extract a model explanation $\CALB$ of a classification model $f$.
    The maximum number of rectangles in each $B_c$ can be controlled by setting $r_{max}$.
    Also, by (\ref{eqn:OptModelExp}) the performance $F_1(\CALB)$ is increasing in $r_{max}$.
    If the value of $r_{max}$ increases, a model explanation has the better performance, while it could be more complex since each $B_c$ has a chance to contain more rectangles.
    On other hand, if the value of $r_{max}$ decreases, the model explanation could be more simple, while the performance $F_1(B_c)$ decreases.
    Consequently, there is a tradeoff between the complexity of a model explanation and its performance, which is controlled by $r_{max}$.


    Before closing this subsection, we would remark the generality of our approach.
    We have derived our method from the viewpoint of the local perturbation that introduced in \cite{ribeiro2016should}.
    However, in our method only a contribution matrix is required to extract the rule-like model explanation, which means the instance explanation step can be replaced with other techniques if we have well-defined interpretable space $\CALF$ and embedding $\varphi$.

\section{Experimental Results} \label{sec:experiments}

    This section provides experimental results of our proposed method to extract a model explanation.
    In our experiments we will use open datasets in the UCI machine learning repository \cite{UCIRepo}.
    Moreover, to show the usability of our method in practice, we will also show the results by an industrial dataset, logs that diagnose the cause of disconnections in a long-term evolution (LTE) network.
    In the dataset, since more than one expert classifies the cause of disconnections, there is a guideline of the categories of causes to ensure the consistency.
    Consequently, we are able to check the validity of our method by comparing the guideline with our results.
    Because of the privacy issue, we will use a filtered dataset with obfuscation.
    So the dataset consists of 10 features whose names are $F_1,\cdots,F_{10}$, and 7 categories $\CALC=\{C_1,\cdots,C_7\}$.

    For our experiments, the random forest (RF), the adaptive boosting (AdaBoost) algorithm and the multi-layer perceptron (MLP) algorithm will be used for classification models.
    We choose these algorithms since they are the typical and basic representatives of ensemble, boosting, and deep learning that are widely used nowadays.
    Our implementation is based on the scikit-learn package \cite{scikit-learn} to train classification models.
    Also, we will generate instance explanations with the help of the LIME \cite{ribeiro2016should}.

\subsection{Validation of Clustering Assumption}

    \begin{figure}[t]
        \centerline{\includegraphics[width=8.0cm]{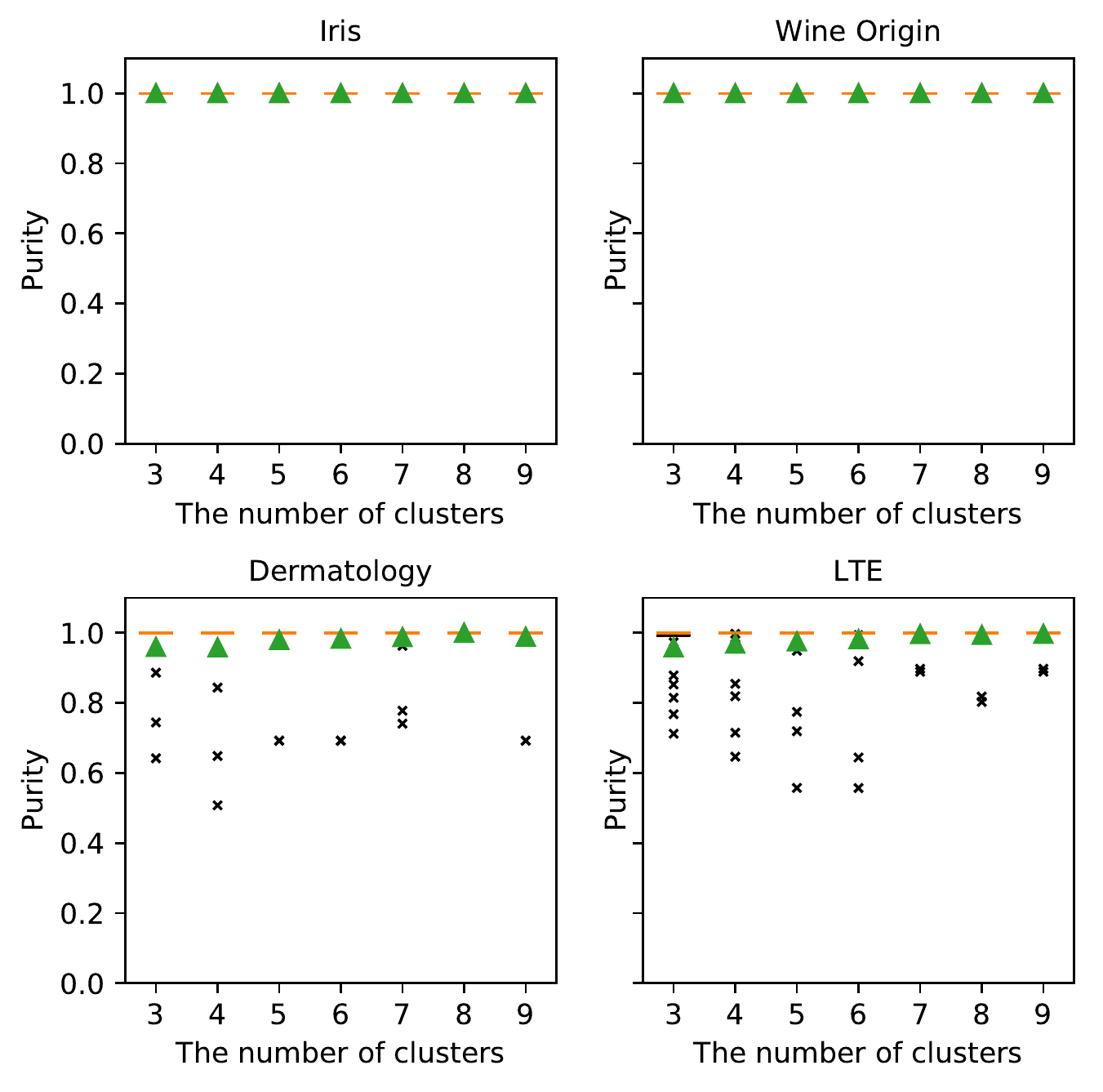}}
        \caption{
            Box-plots of the purity of clusters when the number $r$ of clusters varies.
            The $x$-axis of each plot represents the number $r$ of clusters and the $y$-axis represents the purity of a cluster.
            In each plot, the redline indicates the median value and the triangle marker indicates the average value of the purity.
            Regardless the value of $r$ the value of the purity is almost equal to 1 for all datasets.
            So the box is invisible in all cases.
        }
        \label{fig:purity-vs-r}
    \end{figure}

    Our method has supposed that embedded explanations in the constraint space are well clustered.
    To validate this assumption, we evaluate the purity of each clusters with the K-means clustering.
    The results are plotted in Figure \ref{fig:purity-vs-r}.
    The purity of a cluster is defined by the fraction of the number of embedded explanations in a cluster that are labeled by the most common category in the cluster.
    The $y$-axis of each graph in Figure \ref{fig:purity-vs-r} represents the purity over the clusters.
    The value of the purity is almost 1 in most datasets.
    Even though in the dermatology and LTE cases there are a few clusters having a low purity at $r=3, 4$, both the median or the average are almost 1, which means most clusters have the high purity value.
    Hence, the embedded explanations are well concentrated, which validates our assumption in the previous section.

    \begin{figure}[t]
        \centerline{\includegraphics[width=8.0cm]{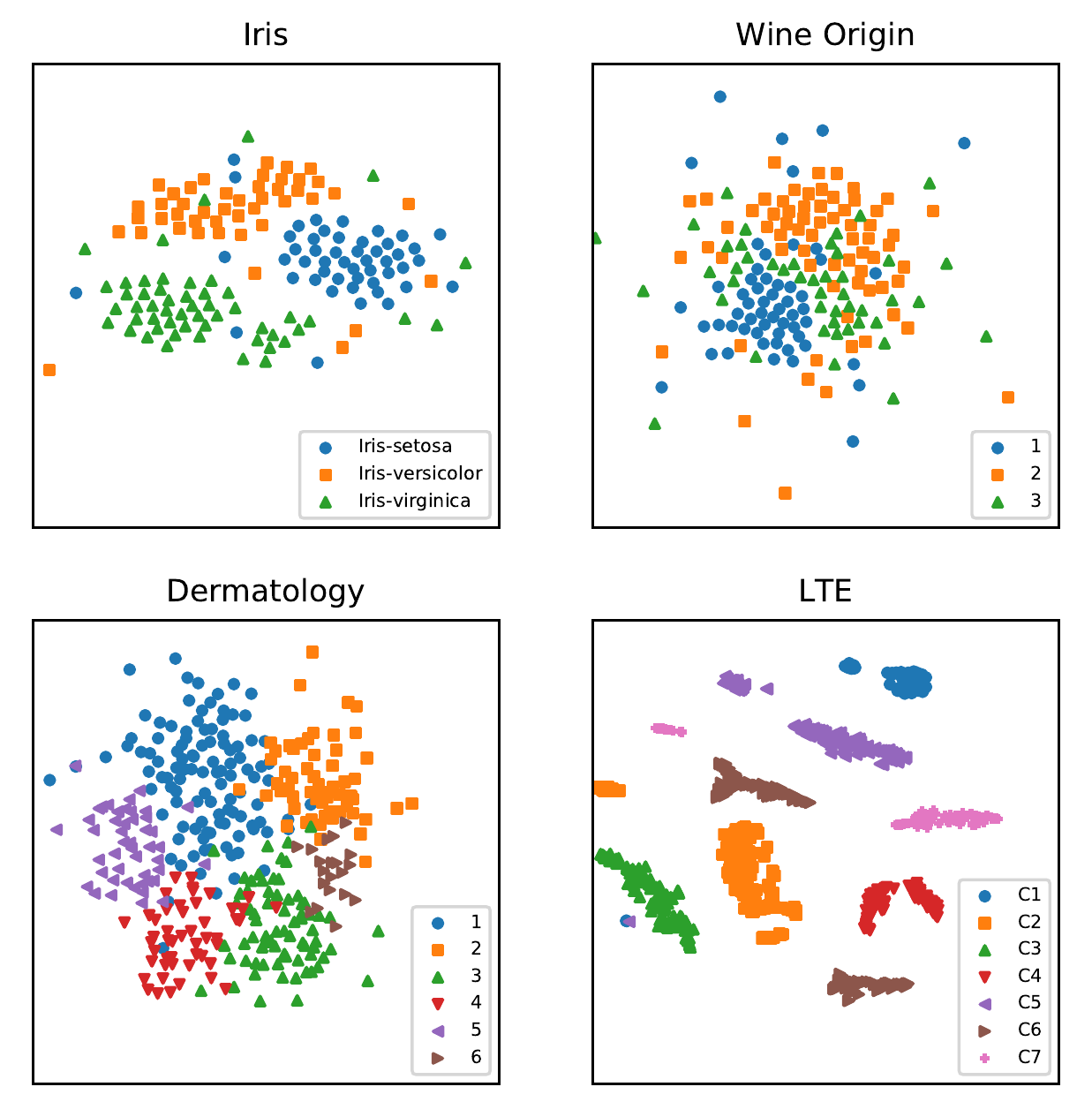}}
        \caption{
            Two dimensional t-SNE of embedded explanations.
            The category of an embedded explanation is indicated by a color.
            In most cases embedded explanations having the same category are well clustered into several groups.
        }
        \label{fig:purity-SNE}
    \end{figure}

    To see in details, we also visualize embedded explanations in Figure \ref{fig:purity-SNE}.
    We visualize the high-dimensional data by using t-SNE \cite{maaten2008visualizing}.
    In fact, t-SNE distills the location of data points to embed into $\BBR^2$-plane, but it is still efficient to investigate the high-dimensional data.
    We use a model that trained by the random forest algorithm.
    Figure \ref{fig:purity-SNE} shows the results of each dataset.
    The category of an instance is indicated by a color.
    In most cases, most embedded explanations having the same category are well clustered into several groups.
    Even the clusters are not clearly separated in the wine origin and dermatology datasets,
    it is still possible to find groups of embedded explanations.

\subsection{Explanation Extraction Results}

    \begin{table}[t]
        \centering
        \caption{
            The performances $F_1(\CALB)$ of model explanation. In the table, we use abbreviations. Wine: Wine origin dataset, Derm: Dermatology dataset, LTE: LTE network dataset.
            A value in the accuracy column is the accuracy of a trained model.
        }
        \begin{tabular}[t]{c|c|c|c|c|c|c}
          \Xhline{3\arrayrulewidth}
          Dataset & $C$ & $m$ & $N$ & Algorithm & Accuracy & $F_1$\\
          \Xhline{3\arrayrulewidth}
          \multirow{3}{*}{Wine}
           & \multirow{3}{*}{3}
           & \multirow{3}{*}{13}
           & \multirow{3}{*}{178}
                 & RF       & 0.98 & 0.92 \\ \cline{5-7}
           & & & & MLP      & 0.97 & 0.94 \\ \cline{5-7}
           & & & & AdaBoost & 0.99 & 0.92 \\
          \hline
          \multirow{3}{*}{Derm}
           & \multirow{3}{*}{6}
           & \multirow{3}{*}{34}
           & \multirow{3}{*}{366}
                  & RF       & 0.98 & 0.77 \\ \cline{5-7}
           &  & & & MLP      & 0.96 & 0.72 \\ \cline{5-7}
           &  & & & AdaBoost & 0.97 & 0.87 \\
          \hline
          \multirow{3}{*}{LTE}
           & \multirow{3}{*}{8}
           & \multirow{3}{*}{10}
           & \multirow{3}{*}{793}
                 & RF       & 0.93 & 0.80 \\ \cline{5-7}
           & & & & MLP      & 0.91 & 0.83 \\ \cline{5-7}
           & & & & AdaBoost & 0.86 & 0.79 \\
          \Xhline{3\arrayrulewidth}
        \end{tabular}
        \label{tbl:moex-results}
    \end{table}

    We now investigate an experiment to see the performance of a model explanation by our method.
    We here fix $k=10$ in the NMF.
    We use 200 decision trees for the RF algorithm.
    The MLP is trained with two hidden layers of 128 nodes and the ReLU activations.
    Finally, the AdaBoost uses the logistic regression as its base estimator.
    In order to train a classification model, we split each dataset into training dataset (70\%) and test dataset (30\%).

    Table \ref{tbl:moex-results} summarizes the performances by our proposed method in terms of $F_1(\CALB)$ defined in (\ref{def:Perf}).
    We see that our algorithm provides the relatively high performance in most cases.
    It means a model explanation $\CALB$ by our method well approximates the partition $\CALR(\CALD)$ by the original model $f$.
    However, the resulting performance in the dermatology dataset has the relatively large variation on the algorithms.
    Due to the simplicity of a model explanation, our method has a limitation on representing an implicit feature like a linear combination of features.
    It also affects the clustering of embedded vectors, which yields that the variance of our method increases over algorithms.
    This issue may be resolved with considering a suitable interpretable space $\CALF$ and an embedding $\varphi$.

    \begin{table}[t]
        \caption{Examples of extracted model explanations.}
        \begin{tabular}[t]{c|l}
          \Xhline{3\arrayrulewidth}
          Category & Model Explanation \\
          \Xhline{3\arrayrulewidth}
          Wine.1 & 12.85 $<$ Alcohol \&  755.0 $<$ Proline \\
          \hline
          \multirow{2}{*}{Wine.2}
                 & (Color intensity $\le$ 3.46  \&  Alcohol $\le$ 12.85) \\
                 & or (Proline $\le$ 682.5) \\
          \hline
          Wine.3 & 3.82 $<$ Color inten. \& Flavanoids $\le$ 0.975 \\
          \Xhline{2\arrayrulewidth}
          \multirow{3}{*}{Derm.1}
                 & (0 $<$ Clubbing of rete ridges) or \\
                 & (1 $<$ Thinning of suprapapillary epidermis \\
                 & \ \& 0 $<$ Elongation of rete ridges) \\
          \hline
          \multirow{5}{*}{Derm.2}
                 & Koebner phenom. $\le$ 0 \& 2 $<$ Spongiosis\\
                 & \ \& Disappearance of granular layer $\le$ 0 \\
                 & \ \& Clubbing of rete ridges $\le$ 0 \\
                 & \ \& Fibrosis of papillary dermis $\le$ 0 \\
                 & \ \& 0 $<$ PNL infiltrate \\
          \Xhline{2\arrayrulewidth}
          LTE.C1 & $0.05649 < F_1$  \\
          \hline
          LTE.C2 & $11 < F_9$ \\
          \hline
          \multirow{2}{*}{LTE.C3}
                 & $4 < F_{10}$ \& $F_1 \le 0.05649$ \& $41 < F_6$ \\
                 & \ \& $F_9 \le 11$ \\
        \Xhline{3\arrayrulewidth}
        \end{tabular}
        \label{tbl:example-moex}
    \end{table}

    Remind that we have interest in the understandability of a model explanation rather than the accuracy.
    To verify the understandability of the results, we provide the examples of extracted model explanations in Table \ref{tbl:example-moex}.
    In the table, due to the space limitation, we denote the category name with its dataset name.
    As seen in the table, the extracted explanation is intuitive.
    For instance, an extracted model explanation of the category {\it setosa} in the iris dataset is given by
    \begin{align*}
        \rm
        petal\ width \le 3,\ petal\ length \le 1.6,\ sepal\ length \le 5.1,
    \end{align*}
    from which we can directly understand that the setosa needs a short length of petal and that of sepal.
    In addition, the results in the LTE network dataset in Table \ref{tbl:example-moex} are almost the same with the guideline written by domain experts.
    Therefore, we conclude that our proposed method can extract a understandable explanation of a classification model.

%

    \begin{figure}[t]
        \centerline{\includegraphics[width=8.0cm]{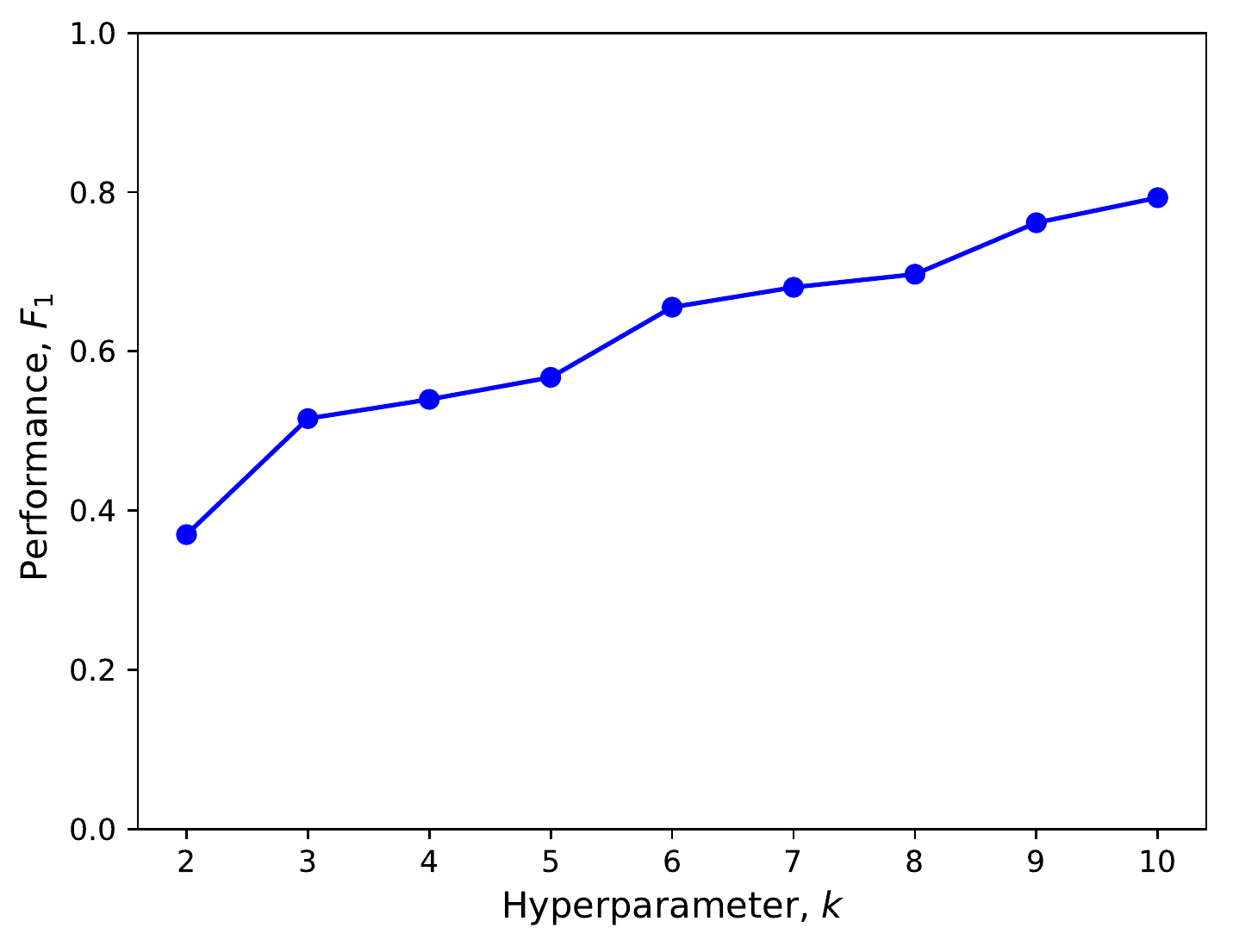}}
        \caption{
            Performance of a model explanation when changing the value of the hyperparameter $k$ in the NMF.
        }
        \label{fig:k_vs_perf}
    \end{figure}

    We next perform experiments to see the effect of the parameter $k$ in the NMF (Figure \ref{fig:k_vs_perf}).
    We use the dermatology dataset since it has the larger number of categories and features than the other dataset has.
    As seen in the figure, the value of $F_1(\CALB)$ increases as the value of $k$ increases.
    The parameter $k$ is related to the number of base constraints that a classification model $f$ has.
    If we have the fewer number of base constraints than the number of categories,
    the base constraints have a weak ability to distinguish categories.
    On the other hand, if $k$ is large enough, base constraints have an ability to distinguish categories separately.
    Consequently, our proposed method needs to have a large enough value of $k$ to extract a proper model explanation, which depends on the number of base constraints of a classification model.

\section{Conclusion}

    In this paper, we propose an analytic method that extracts a understandable model explanation of a black-box classification model by using the nonnegative matrix factorization.
    We consider a concept of a model explanation in terms of the set approximation.
    We introduce a contribution matrix of features based on an interpretable explanation.
    By applying the NMF to the contribution matrix we extract the underlying decision making constraints of the classification model.
    Then combining those constraints we find a model explanation of the classification model.
    To see the validity of our approach, experiments are performed with open datasets as well as an industrial dataset.
    The experimental results show that our method extracts understandable and reasonable rule-like explanation of the model.
    So our method can help a service provider quickly decide the validity of a black-box model,
    and it can also help in reducing the development costs of a new intelligent service in practice.

\bibliography{ModelExplanation}

\begin{thebibliography}{}

\bibitem[\protect\citeauthoryear{Bastani, Kim, and
  Bastani}{2017}]{bastani2017interpreting}
Bastani, O.; Kim, C.; and Bastani, H.
\newblock 2017.
\newblock Interpreting blackbox models via model extraction.
\newblock {\em arXiv:1705.08504}.

\bibitem[\protect\citeauthoryear{Fong and
  Vedaldi}{2017}]{fong2017interpretable}
Fong, R., and Vedaldi, A.
\newblock 2017.
\newblock Interpretable explanations of black boxes by meaningful perturbation.
\newblock {\em arXiv:1704.03296}.

\bibitem[\protect\citeauthoryear{Kim, Rudin, and Shah}{2014}]{kim2014bayesian}
Kim, B.; Rudin, C.; and Shah, J.~A.
\newblock 2014.
\newblock The bayesian case model: A generative approach for case-based
  reasoning and prototype classification.
\newblock In {\em Advances in Neural Information Processing Systems},
  1952--1960.

\bibitem[\protect\citeauthoryear{Kotsiantis, Zaharakis, and
  Pintelas}{2007}]{kotsiantis2007supervised}
Kotsiantis, S.~B.; Zaharakis, I.; and Pintelas, P.
\newblock 2007.
\newblock Supervised machine learning: A review of classification techniques.

\bibitem[\protect\citeauthoryear{Li and Belford}{2002}]{Li2002}
Li, R.-H., and Belford, G.~G.
\newblock 2002.
\newblock Instability of decision tree classification algorithms.
\newblock In {\em Proceedings of the Eighth ACM SIGKDD International Conference
  on Knowledge Discovery and Data Mining}, KDD '02,  570--575.
\newblock New York, NY, USA: ACM.

\bibitem[\protect\citeauthoryear{Lichman}{2013}]{UCIRepo}
Lichman, M.
\newblock 2013.
\newblock {UCI} machine learning repository.

\bibitem[\protect\citeauthoryear{Lipton}{2016}]{lipton2016mythos}
Lipton, Z.~C.
\newblock 2016.
\newblock The mythos of model interpretability.
\newblock In {\em ICML Workshop on Human Interpretability of Machine Learning}.

\bibitem[\protect\citeauthoryear{Maaten and
  Hinton}{2008}]{maaten2008visualizing}
Maaten, L. v.~d., and Hinton, G.
\newblock 2008.
\newblock Visualizing data using t-sne.
\newblock {\em Journal of Machine Learning Research} 9(Nov):2579--2605.

\bibitem[\protect\citeauthoryear{N{\'u}{\~n}ez, Angulo, and
  Catal{\`a}}{2002}]{nunez2002rule}
N{\'u}{\~n}ez, H.; Angulo, C.; and Catal{\`a}, A.
\newblock 2002.
\newblock Rule extraction from support vector machines.
\newblock In {\em Esann},  107--112.

\bibitem[\protect\citeauthoryear{Pedregosa \bgroup et al\mbox.\egroup
  }{2011}]{scikit-learn}
Pedregosa, F.; Varoquaux, G.; Gramfort, A.; Michel, V.; Thirion, B.; Grisel,
  O.; Blondel, M.; Prettenhofer, P.; Weiss, R.; Dubourg, V.; Vanderplas, J.;
  Passos, A.; Cournapeau, D.; Brucher, M.; Perrot, M.; and Duchesnay, E.
\newblock 2011.
\newblock Scikit-learn: Machine learning in {P}ython.
\newblock {\em Journal of Machine Learning Research} 12:2825--2830.

\bibitem[\protect\citeauthoryear{Ribeiro, Singh, and
  Guestrin}{2016}]{ribeiro2016should}
Ribeiro, M.~T.; Singh, S.; and Guestrin, C.
\newblock 2016.
\newblock Why should i trust you?: Explaining the predictions of any
  classifier.
\newblock In {\em Proceedings of the 22nd ACM SIGKDD International Conference
  on Knowledge Discovery and Data Mining},  1135--1144.
\newblock ACM.

\bibitem[\protect\citeauthoryear{Ross, Hughes, and
  Doshi-Velez}{2017}]{RossHD17}
Ross, A.~S.; Hughes, M.~C.; and Doshi-Velez, F.
\newblock 2017.
\newblock Right for the right reasons: Training differentiable models by
  constraining their explanations.
\newblock {\em arXiv:1703.03717}.

\bibitem[\protect\citeauthoryear{Shrikumar, Greenside, and
  Kundaje}{2017}]{ShrikumarGK17}
Shrikumar, A.; Greenside, P.; and Kundaje, A.
\newblock 2017.
\newblock Learning important features through propagating activation
  differences.
\newblock {\em arXiv:1704.02685}.

\bibitem[\protect\citeauthoryear{Simonyan, Vedaldi, and
  Zisserman}{2014}]{SimonyanVZ13}
Simonyan, K.; Vedaldi, A.; and Zisserman, A.
\newblock 2014.
\newblock Deep inside convolutional networks: Visualising image classification
  models and saliency maps.
\newblock {\em 2nd International Conference on Learning Representations
  Workshop}.

\bibitem[\protect\citeauthoryear{Turner}{2016}]{Turner2016}
Turner, R.
\newblock 2016.
\newblock A model explanation system.
\newblock In {\em 2016 IEEE 26th International Workshop on Machine Learning for
  Signal Processing (MLSP)},  1--6.

\bibitem[\protect\citeauthoryear{Witten \bgroup et al\mbox.\egroup
  }{2016}]{witten2016data}
Witten, I.~H.; Frank, E.; Hall, M.~A.; and Pal, C.~J.
\newblock 2016.
\newblock {\em Data Mining: Practical machine learning tools and techniques}.
\newblock Morgan Kaufmann.

\bibitem[\protect\citeauthoryear{Zintgraf \bgroup et al\mbox.\egroup
  }{2017}]{ZintgrafCAW17}
Zintgraf, L.~M.; Cohen, T.~S.; Adel, T.; and Welling, M.
\newblock 2017.
\newblock Visualizing deep neural network decisions: Prediction difference
  analysis.
\newblock {\em arXiv:1702.04595}.

\end{thebibliography}
\bibliographystyle{aaai}

\end{document}